\documentclass[colorlinks,upint,subscriptcorrection,varvw,hyphenate,balance,USenglish]{asmeconf_HAL} 

\hypersetup{%
	pdfauthor={Léonie Plancoulaine},									  
	pdftitle={Design and stability analysis of an underactuated hand with passively rotating fingers},                  
	pdfkeywords={Underactuated robotic hand, gripping dexterity, stability, spatial mobility, passivity},
	pdfsubject = {Underactuated robotic hand},			  
}

\allowdisplaybreaks 

\usepackage[utf8]{inputenc}
\usepackage[T1]{fontenc}
\usepackage{algorithmic}
\usepackage{graphicx}
\usepackage{textcomp}
\usepackage{xcolor}
\usepackage{booktabs}
\usepackage{threeparttable}
\usepackage{tabularx}

\usepackage{xcolor}

\begin{document}

\ConfName{}
\ConfAcronym{}
\ConfDate{} 
\ConfCity{}
\PaperNo{}

\title{Design and stability analysis of an underactuated hand with passively rotating fingers}
\SetAuthors{%
    Léonie Plancoulaine\affil{1}, 
	Sylvain Guegan\affil{2}, 
    Franck Plestan\affil{1},
	Damien Chablat\affil{1}\CorrespondingAuthor{damien.chablat@cnrs.fr}
	}

\SetAffiliation{1}{Nantes Université, École Centrale Nantes, CNRS, LS2N, UMR 6004,  44300 Nantes, France}
\SetAffiliation{2}{Université Rennes, INSA Rennes, LGCGM, F-35000 Rennes, France}

\maketitle

\begin{abstract}
This paper presents an innovative design and stability analysis of an underactuated robotic finger with spatial mobility, designed to enhance gripping dexterity in robotic hands. The finger architecture incorporates a revolute joint at its base, enabling passive spatial rotation that facilitates both cylindrical and spherical grasping. With only two phalanges per finger, the design simplifies kinematic complexity while supporting precision and enveloping grasps. Stability criteria, based on the moment at the finger base joint induced by contact forces, are introduced to ensure reliable object gripping and prevent ejection during manipulation. The study also examines a differential mechanism that distributes a single actuation torque across multiple fingers, allowing adaptive and coordinated motion. This mechanism enhances the hand’s ability to grasp diverse object shapes with minimal pre-grasp adjustments, leveraging passivity for autonomous adaptation. Theoretical findings are experimentally validated using a fully mechanical prototype, demonstrating versatility in performing cylindrical, spherical, parallel, and enveloping grasps. The integration of underactuation—both within individual fingers and among multiple fingers—reduces mechanical complexity, cost, and control demands while preserving functional adaptability. This work advances the development of compliant robotic hands suitable for applications requiring dexterity and robustness, such as agricultural robotics, logistics, assistive technologies, and waste sorting. Future research will focus on automating actuation and refining control strategies to further improve grasp stability and precision, paving the way for autonomous manipulation in unstructured environments.
\end{abstract}

\section*{Notation}
In the following, the sine and cosine functions are abbreviated as 
$s \cdot = \sin\cdot$ and $c \cdot = \cos\cdot$. 
For instance, $c\psi = \cos\psi$ and $s\psi = \sin\psi$.
\section{Introduction}
The human hand is a remarkably complex and dexterous tool, capable of executing a wide variety of grasping strategies. According to Feix's taxonomy~\cite{feix_grasp_2016}, humans can perform up to 34 distinct types of grasps, depending not only on the shape of the object but also on the intended task. While robotic grippers are commonly used for repetitive tasks in industrial settings, they fall short of replicating the agility and adaptability of the human hand.

The study of compliant robotic hands is particularly relevant for applications such as bulk separation, harvesting, packaging, waste sorting, and warehouse logistics \cite{piazza_century_2019}. Dexterity, as defined by Seguin~\cite{seguin_approach_2023}, refers to the ability to maintain a stable object configuration while moving it in-hand from an initial position to a final one.
Dexterity can be divided into two key components: grasping and in-hand manipulation. This work focuses specifically on grasping.

Although two fingers are sufficient for basic grasping tasks, this configuration limits both dexterity and stability~\cite{pons_multifingered_1999}. A three-fingered design offers improved grasping capabilities while maintaining mechanical simplicity. Various types of robotic hands have been developed~\cite{piazza_century_2019,kadalagere_sampath_review_2023,ramirez_rebollo_3_2017}, including soft grippers, fully actuated hands, and underactuated hands.

Soft grippers~\cite{lee_soft_2017,shintake_soft_2018}, made from compliant materials, offer excellent adaptability for handling delicate objects, but suffer from limitations in force control, precision, and durability. Fully actuated hands~\cite{butterfass_dlr-hand_2001,jacobsen_design_1986}, which include one actuator per degree of freedom, provide high precision but are costly and complex to control. 

In \cite{Backus_2016}, the robotic hand features one actuated prismatic joint and one passive rotational joint per finger, enabling adaptive grasping of diverse object shapes. The hand's adaptive grasping is limited by the fixed positioning of the passive rotational joint axis, which can result in suboptimal contact and reduced grasp stability for very small or irregularly shaped objects.

Underactuated hands ~\cite{gosselin_underactuation_2017,birglen_underactuated_2008}, by contrast, use fewer actuators than degrees of freedom, relying on passive elements such as springs to manage the remaining joints. This approach simplifies the system, reduces cost, and enables shape adaptation, albeit with reduced precision. This paper focuses on underactuated hands to achieve a balance between adaptability, simplicity, and robustness. Underactuation can be implemented either within individual fingers or between multiple fingers~\cite{birglen_underactuated_2008}. 

In the case of underactuated fingers, the motion of each phalanx depends on the others, directly influencing the types of grasp that can be achieved. Three primary grasping modes  are considered in the literature~\cite{dang_cosa-let_2017}:
\begin{itemize}
    \item \textbf{Coupled grasp}: Phalanges are linked by a linear relationship between their rotation angles, mimicking the natural pre-shaping of the human hand.
    \item \textbf{Parallel grasp}: The distal phalanx remains parallel to its initial orientation, enabling precision grasps.
    \item \textbf{Self-adaptive grasp}: The finger remains straight until the proximal phalanx contacts the object. Once contact is made, and the actuation torque exceeds the spring resistance, the distal phalanx rotates until it also makes contact.
\end{itemize}

Hybrid grasping strategies are also possible. For instance, a finger may begin in coupled mode and switch to self-adaptive once the proximal phalanx is blocked. Similarly, parallel and self-adaptive modes can be combined to enable both precision and enveloping grasps.

Most underactuated fingers operate in a planar configuration. However, some designs incorporate additional degrees of freedom, such as abduction/adduction~\cite{gosselin_underactuation_nodate-3}, or passive spatial mobility~\cite{hamon2023model}. Although these enhancements increase grasp diversity, they often come at the cost of added kinematic complexity and reduced precision. In contrast, the proposed finger design integrates a passive revolute joint at its base, enabling spatial mobility while maintaining a simplified kinematic structure. 

About underactuation among multiple fingers, several types of differential systems can be used \cite{birglen_underactuation_2008}. In such a configuration, a single actuator can drive the opening and closing of all fingers while adding adaptability. If one or more fingers are blocked, the others can continue to move and compensate. Once all fingers are in contact, the applied force is distributed among them to achieve a stable grasp. Different differential mechanisms can be implemented to realize this principle, including pulleys, seesaw mechanisms, gear trains, or fluidic systems. In this work, following the approach of \cite{massa_design_2002}, a spring-loaded slider is adapted. This solution is compact and allows for an arbitrary number of outputs.

This paper presents a simplified kinematic design that enables passive self-rotation of each finger, allowing the hand to perform a wide range of grasps with a single actuator. The following sections describe the finger and hand architecture, introduce a stability analysis for parallel and self-adaptive grasps, examine the force distribution across the three outputs of the differential mechanism, present experimental validation using a physical prototype, and conclude with a discussion and future research directions.

\section{Hand architecture}
\label{sec:hand architecure}
This section presents the hand\footnote{Animation of the different levels of underactuation: https://youtu.be/29pfHB39d48} architecture developed by the authors. Its purpose is to use a single actuator to minimize pre-grasp adjustments while enabling the grasping of objects of various shapes through embodied intelligence, by using the mechanism’s passivity to autonomously adapt. This passivity is used both at the finger level and in their coordinated motion.

\subsection{Finger architecture}
\begin{figure}[!ht]
    \centering
    \includegraphics[width=0.6\linewidth]{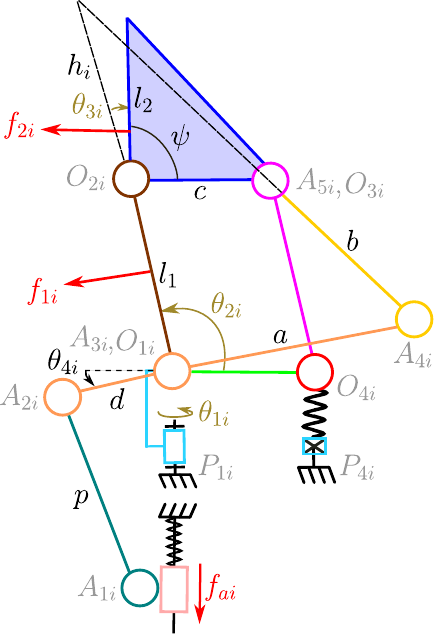}
    \caption{Two-dimensional kinematic model of a finger with two interlinked closed loops, providing three DOF through a single actuation at $A_{1i}$ and underactuation via $S_{2i}$.}
    \label{fig:Finkin}
\end{figure}

\begin{figure}[!ht]
\centerline{\includegraphics[width=0.6\linewidth]{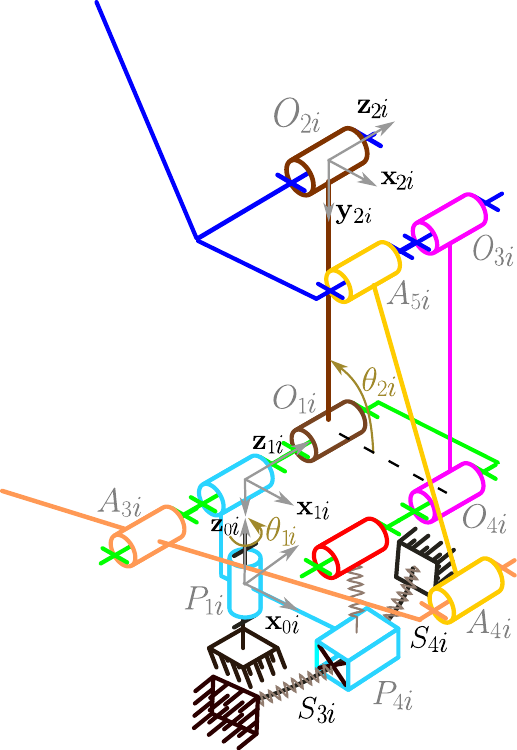}}
\caption{Three-dimensional kinematic model of a finger with a passive rotation at $P_{1i}$ through springs $S_{3i}$ and $S_{4i}$.}
\label{fig:cinema}
\end{figure}

The kinematic structure of the finger is illustrated in Figure~\ref{fig:Finkin}, which shows the detailed kinematic configuration of the finger, and Fig.~\ref{fig:cinema}, which presents the kinematics model highlighting the dynamic interactions within the system. The parameters introduced here are generic for the three fingers, where $i$ denotes the $i^\text{th}$ finger. 

\begin{figure}
\centerline{\includegraphics[width=0.30\linewidth]{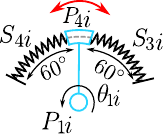}}
\caption{Top view of the revolute joint at the finger base $P_{1i}$ for the finger passive rotation.}
\label{fig:rotation}
\end{figure}

To enable spatial motion, a revolute joint is introduced at the base of each finger (point $P_{1i}$), Fig.~\ref{fig:rotation}. This joint allows the finger to operate in a three-dimensional space rather than being limited to planar motions without increasing mechanical complexity. Antagonist springs $S_{3i}$ and $S_{4i}$ ensure the finger returns to its initial position after the grasping process.

The finger's kinematic behavior is best understood by examining its structure, which is composed of two interlinked closed loops. Specifically, the actuation loop ($A_{3i}$, $A_{4i}$, $A_{5i}$, $O_{2i}$) drives the motion, while the parallel grasp loop ($O_{1i}$, $O_{4i}$, $O_{3i}$, $O_{2i}$) enables both parallel and enveloping grasp capabilities.  For the switching between the two modes, spring $S_{2i}$ plays a key role: its extension drives the movement of the distal phalanx.
For simplicity, each finger is modeled with two phalanges.
The finger possesses three degrees of freedom (DOF) (Fig.~\ref{fig:Finkin}), defined as follows:
\begin{itemize}
\item $\theta_{1i}$, angle between axis ${\bf x}_{0i}$ and ${[P_{1i}P_{4i}]}$ along $\bf{z}_{0i}$; 
\item $\theta_{2i}$, angle between axis ${\bf x}_{1i}$ and the proximal phalanx along $\bf{z}_{1i}$;
\item $\theta_{3i}$, angle between the proximal and distal phalanges along $\bf{z}_{2i}$.
\end{itemize}
The contact forces $f_{1i}$ and $f_{2i}$ act on the proximal and distal phalanges, respectively. $f_{ai}$ represents the finger actuation effort. The angle $\theta_{4i}$ defines the angle between axis $\bf{x}_{0i}$ and ${[A_{2i}A_{3i}]}$.
Different dimensions define the finger kinematics: 
\begin{itemize}
\item $a$, the distance between $A_{3i}$ and $A_{4i}$;
\item $b$, the distance between $A_{5i}$ and $A_{4i}$;
\item $c$, the distance between $O_{2i}$ and $O_{3i}$;
\item $d$, the distance between $A_{2i}$ and $A_{3i}$;
\item $p$, the distance between $A_{1i}$ and $A_{2i}$;
\item $l_1$, the length of the proximal phalanx;
\item $l_2$, the length of the distal phalanx;
\item $\psi$, constant angle between the distal phalanx and ${[O_{2i}O_{3i}]}$.
\end{itemize}
\subsection{Differential mechanism}
While two fingers suffice for basic grasping tasks in structured environments, they lack the versatility required for diverse object manipulation. Three fingers offer improved dexterity, and four enable in-hand manipulation~\cite{pons_multifingered_1999}. This study focuses on grasping with three fingers.

\begin{figure}[!ht]
\centerline{\includegraphics[width=1\linewidth]{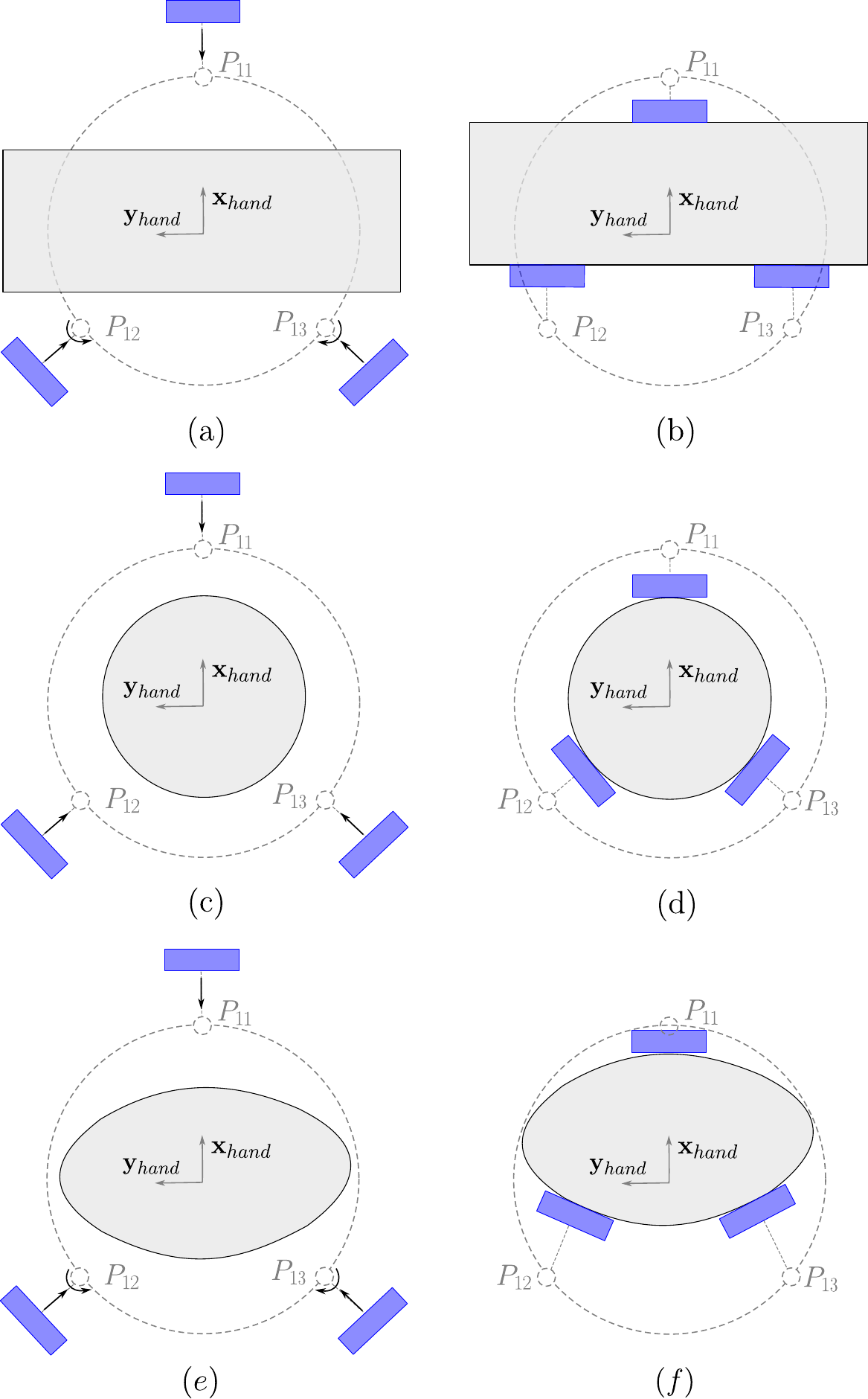}}
\caption{Hand performing a cylindrical and a spherical grasp viewed from
above: (a)(c)(e) initial fingers position (b) passive adaptation of the fingers for a cylindrical grasp (d) passive adaptation of the fingers for a spherical grasp and (f) passive adaptation for an intermediate grasp.}
\label{fig:position}
\end{figure}

To illustrate the grasping configuration, Fig.~\ref{fig:position} shows cylindrical and spherical grasps: cylindrical grasps occur when two fingers oppose a third, while spherical
grasps arise when all three fingers converge toward a common point \cite{miller_automatic_2003}. Intermediate grasps occur when the finger configuration is between cylindrical and spherical patterns. Each finger is positioned 120° apart and is capable of rotating 120° around its own axis. This design enables any of the three fingers to act as a thumb, significantly enhancing the hand's versatility. The hand is able to grasp both simple and asymmetrical objects, thanks to this passivity.

\begin{figure} [!ht]
    \centering
    \includegraphics[width=0.85\linewidth]{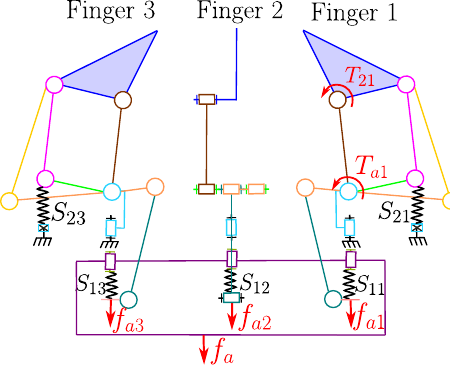}
    \caption{The differential mechanism with a global actuation force $f_a$ and three output forces $f_{a1}$, $f_{a2}$ and $f_{a3}$.}
    \label{fig:diffmec}
\end{figure}

The differential mechanism proposed in this work, Fig.~\ref{fig:diffmec}, is composed of three springs. It incorporates a sliding element actuated manually, which can later be driven by an electric motor or a pneumatic actuator. The sliding motion compresses the three springs. If the movement of one finger is blocked, the corresponding spring compensates by absorbing the displacement, thereby allowing the other two fingers to continue moving. Unlike the slider proposed by Massa \cite{massa_design_2002}, the one presented here is connected to mechanical linkages, enabling torque transmission. It is composed of two revolute joints that provide a lever-like motion.\\

\section{Stability analysis}
This section addresses the stability analysis of parallel and enveloping grasps, with particular attention to finger rotation. In both theoretical grasp analyses, the force is considered as a point force. Birglen \cite{birglen_grasp-state_2006} states that assuming the force is a single point contact is the grasp stability worst-case scenario. In this study, the spring effects of $S_{3i}$ and $S_{4i}$ are neglected to focus on the analysis of contact forces. However, their stiffness directly influences the finger’s rotation, driving it back to its initial position. The finger stability study results are generic and valid for all three fingers. Furthermore, it is essential to examine how the torques are distributed among the three outputs of the differential mechanism. 

\subsection{Precision grasp}
To analyze the mechanical behavior of the finger, it is treated as an isolated system. For this study, the finger weight is considered negligible. The joint angle $\theta_3$ is assumed to be equal to $\theta_{2i}-\psi$ as the distal phalanx remains parallel to its initial orientation during precision grasps. 

\begin{figure}[!ht]
    \centering
    \includegraphics[width=0.55\linewidth]{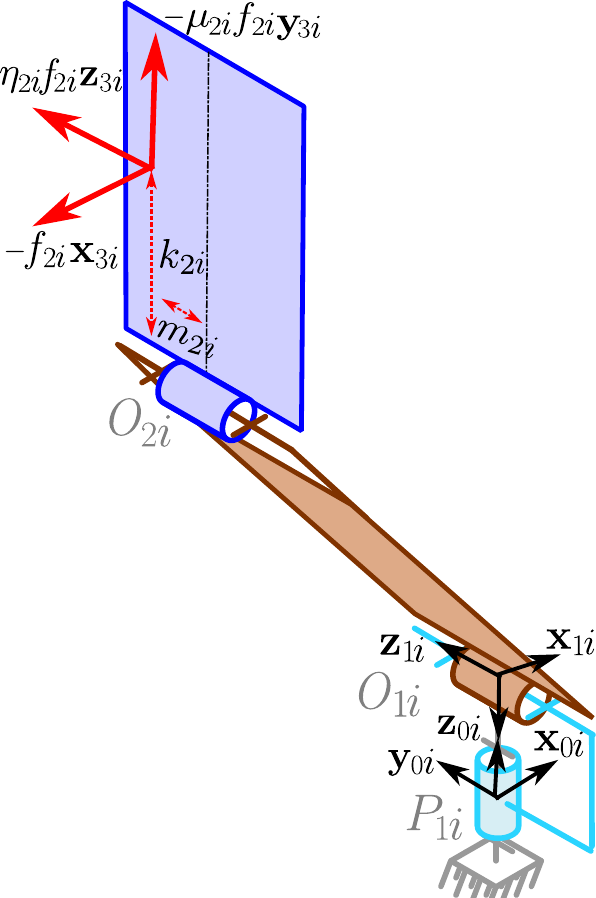}
    \caption{Illustration of the contact force and its location used in precision grasp.}
    \label{fig:precision_grasp}
\end{figure}

After isolating the finger, a force analysis is performed. Forces and torques (i.e. mechanical actions due to the actuation, revolute joint ($P_{1i}$) and contact forces) are expressed in $R_{0i}$, ($P_{1i}$,$\bf{x}_{0i}$,$\bf{y}_{0i}$,$\bf{z}_{0i}$) the base frame of finger i attached to $P_{1i}$. 
The actuation force is applied manually and generates a torque around the joint at point $O_{1i}$. The spring contributes to the orientation of the distal phalanx according to its stiffness, thereby affecting the finger’s passive behavior. For precision grasp, the spring remains nearly unstretched, generating negligible torque. The revolute joint at point $P_{1i}$ introduces reaction forces due to joint constraints. Finally, the contact force from the object is decomposed into components that depend on the object-phalanx friction coefficients $\mu_{1i}$ and $\eta_{1i}$, as well as on the geometric parameters $l_1$, $\psi$, the position of the contact point $k_{2i}$, $m_{2i}$, and the joint angles $\theta_{1i}$ and $\theta_{2i}$. These expressions are essential for evaluating grasp stability and understanding how each mechanical element contributes to the overall equilibrium of the system. Figure~\ref{fig:precision_grasp} illustrates the relevant geometric and frictional parameters used in the stability analysis. 

The sign of the moment $M_{z}$ about the  ${\bf z}_0$ axis must be evaluated to determine whether the contact configuration leads to a stable grasp. By applying the fundamental principle of statics, it appears that $M_z$ depends only on the contact force, leading to the following result:
\begin{equation}
    M_{z} = f_{2i}((k_{2i} \eta_{2i}-m_{2i} \mu_{2i}) c\psi+ l_1 \eta_{2i} c\theta_{2i} + m_{2i} s\psi).
    \label{moment1}
\end{equation}

In this expression:
\begin{itemize}
    \item $m_{2i}$ is the distance along ${\bf z}_{3i}$ between the revolute joint at $O_{2i}$ and the contact point of $f_{2i}$,
    \item $k_{2i}$ is the distance along ${\bf y}_{3i}$ between the same joint and the contact point of $f_{2i}$,
    \item $\mu_{2i}$ and $\eta_{2i}$ are friction coefficients defined by $f_{t2i} = \mu_{2i} f_{2i}$ and $\tau_{2i}= \eta_{2i} f_{2i}$, with bounds $-\mu_{\text{static}} \leq \mu_{2i} \leq \mu_{\text{static}}$ and $-\eta_{\text{static}} \leq \eta_{2i} \leq \eta_{\text{static}}$.
\end{itemize} 
Additional parameters such as $f_{2i}$, $l_1$, and the angle $\psi$ are defined in Section~\ref{sec:hand architecure}.

To ensure stable contact between the object and the phalanx, the contact force must remain within the previously defined physical boundaries of the phalanx, and the finger must rotate toward a stable position along the phalanx to prevent object ejection. If these conditions are not met, the object may be ejected due to improper force orientation or unstable contact geometry. 

\subsection{Power grasp}
\label{sec:power grasp}
Birglen \cite{birglen_design_2008} defines a quasi-static stability condition for an enveloping grasp. This criterion states that a grasp is stable if and only if the finger is in static equilibrium, meaning the contact forces between the object and the phalanges are non-negative. An underactuated finger cannot guarantee that all phalanges remain in contact with the object during an enveloping grasp. Since there are fewer actuators than degrees of freedom, the grasp force distribution and stability depend on the mechanical design of the hand.

\begin{figure}[!ht]
    \centering
    \includegraphics[width=0.55\linewidth]{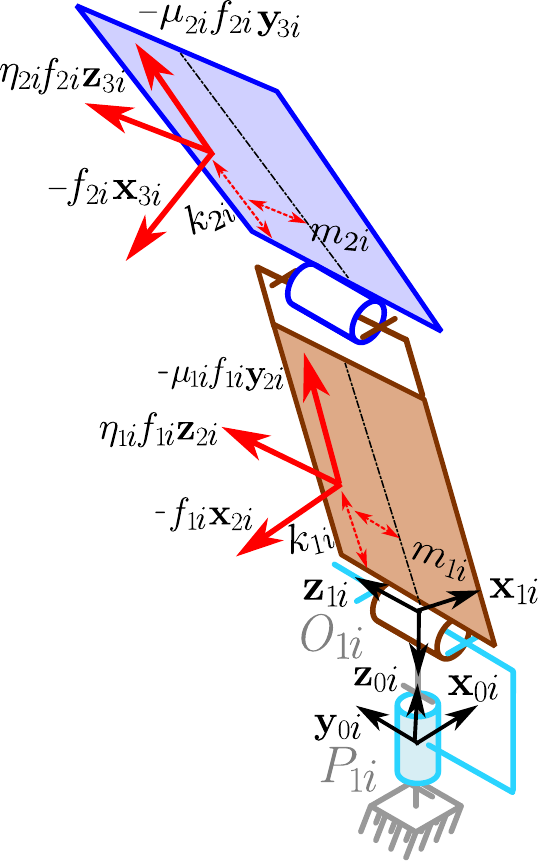}
    \caption{Illustration of the contact forces and their location used in enveloping grasp.}
    \label{fig:enveloppant}
\end{figure}
The vector of contact forces $\bf{f}$ can be expressed as a function of \cite{birglen_design_2008}: ${\bf J}$, the Jacobian matrix that depends on the position of the contact point on the phalanx, the relative orientation of the phalanges, and the friction coefficients, ${\bf T}$, the transmission matrix, determined by the mechanical transmission of the actuator torque, and ${\bf t}$, the vector of input torques generated by the actuator and springs. The contact forces are given by:
\begin{equation}
{\bf f} = {\bf J}^{-t}{\bf T}^{-t} {\bf t}.
\end{equation}

The mechanism consists of a serial chain with three revolute joints. Unlike Birglen's design \cite{birglen_design_2008}, this mechanism allows for spatial movement. The initial rotation, and thus $\theta_1$, does not affect the force distribution among the phalanges, as it does not change the kinematic configuration of the enveloping grasp.

Birglen's result in the plane can then be used for the stability study:

\begin{equation}
{\bf f_i} =
\begin{bmatrix}
\dfrac{-T_{2i} (h_i + l_{1})A(B + C)+ T_{ai} (h_i A(B+C) -(h_i + l_{1})D)}{(h_i + l_{1})AD}\\
\dfrac{T_{2i}(h_i + l_{1}) + T_{ai} h_i}{(h_i + l_{1}) B}
\label{forcecontact}
\end{bmatrix}
\end{equation}

where:
\begin{equation*}
A = k_{1i} -\eta_{1i} 
\end{equation*}
\begin{equation*}
B = k_{2i}-\eta_{2i}
\end{equation*}
\begin{equation*}
C = l_1(c\theta_{3i}+\mu_{2i} s\theta_{3i})
\end{equation*}
\begin{align*}
D =(\eta_{1i}-k_{1i})(\eta_{2i}-k_{2i})
 \\
\end{align*}

$T_{ai}$ is the actuation torque at point $O_{1i}$ and $T_{2i}$ is the torque at $O_{2i}$. $m_{1i}$ and $k_{1i}$ are, respectively, the distance along $\bf{z}_{2i}$ and $\bf{y}_{2i}$ between the revolute joint at $O_{1i}$ and the contact point of $f_{1i}$. $h_i$ is the signed distance between $O_{2i}$ and the intersection of lines $(O_{1i}O_{2i})$ and $(A_{4i}A_{5i})$.

Thanks to the force distribution, one can study the moment around $\bf{z}_{0i}$ to ensure that the finger rotates in the right direction and to study the friction influence.
By applying the fundamental principle of statics:

\begin{align}
M_{z} &= f_{2i}\Big((k_{2i} \eta_{2i} - m_{2i} \mu_{2i})c(\theta_{2i}+\theta_{3i})\notag \\
& \quad + l_1 \eta_{2i} c\theta_{2i} 
        - m_{2i} s(\theta_{2i}+\theta_{3i})\Big) \notag\\
      &\quad + f_{1i}\Big((k_{1i}\eta_{1i} - m_{1i} \mu_{1i}) c\theta_{2i} + m_{1i} s\theta_{2i}\Big).
\label{moment}
\end{align}

\subsection{Distribution of the actuation torque}
Due to the differential mechanism, the actuation torque of each finger is not identical. It is therefore important to understand the factors influencing it.

Considering the slider in isolation, the forces acting on it can be expressed as: ${f_a}$, the input actuation force, and ${f_{a1}}$, ${f_{a2}}$, and ${f_{a3}}$, the output forces of the differential mechanism.

\begin{equation}
\begin{bmatrix}
f_{a1} \\
f_{a2} \\
f_{a3}
\end{bmatrix}
=
\begin{bmatrix}
1 & 0 & -1&-1 \\
1 &  -1 &  0&-1 \\
1 &  -1 & - 1&0
\end{bmatrix}
\begin{bmatrix}
f_{a} \\
f_{s1}\\
f_{s2} \\
f_{s3}
\end{bmatrix}.
\label{eq:distrib}
\end{equation}

where $f_{si}=K(x_{si}-x_0)$ with $x_0$ the spring free length and $x_{si}$  the $i^{\text{th}}$ finger spring length at the final position.

The actuation torque of each finger is generated by the actuation force transmitted through the mechanism. To determine the torque of the $i^{\text{th}}$ finger, $T_{ai}$, we compute the moment at point $A_{3i}$. Since the actuation force $\mathbf{F}_{Ai}$ applied at $A_{1i}$ is collinear with $\mathbf{A_{1i}A_{3i}}$, it does not contribute directly to the torque. Instead, the reaction force at the revolute joint $A_{2i}$, denoted $\mathbf{R}_{A_{2i}}$, generates the actuation torque at $A_{3i}$.

Assuming quasi-static motion and applying the principle of static equilibrium:

\begin{equation}
\mathbf{M_{A3i}} 
= \mathbf{A_{3i}A_{2i}} \times \mathbf{R}_{A_{2i}} 
=
\begin{pmatrix}
0 \\[1mm]
f_{ai} \, d \, c\theta_{4i} \\[1mm]
0
\end{pmatrix}.
\end{equation}

Hence, the actuation torque is given by:
\begin{equation}
T_{ai} = f_{ai}  d  c\theta_{4i}
\end{equation}
with:
\begin{equation}
\theta_{4i} = -\frac{\pi}{2}+\arccos{\left(\frac{p^2+d^2-(x_g+x_{si})^2}{2pd}\right)}
\end{equation}
where $x_g$ is the distance between the top of the slider and $A_{3i}$ on $\bf {z}_{0i}$.
\section{Results}
The geometric parameters listed in Table~\ref{tab:geometric_parameters} are used throughout the different studies, with the angles $\theta_{1i}$, $\theta_{2i}$, and $\theta_{3i}$ varying within the ranges $-60^\circ < \theta_{1i} < 60^\circ$, $-140^\circ < \theta_{2i} < -52^\circ$, and $-65^\circ < \theta_{3i} < 45^\circ$, respectively. Geometric parameter values are optimized using Birglen’s study \cite{birglen2004optimal}.

\begin{table}[!ht]
\caption{Geometric parameters of the fingers}
\centering
\renewcommand{\arraystretch}{1.3}
\begin{tabularx}{\columnwidth}{l l l l l l l}
\toprule
$l_1$\footnotesize{(mm)} & $l_2$\footnotesize{(mm)} &$n_{1,2}$\footnotesize{(mm)} & $\psi$(°) & $a$\footnotesize{(mm)}&$b$\footnotesize{(mm)}&$c$\footnotesize{(mm)}\\
\midrule
50.7  & 38 &31.9 &97&33.48&54&15\\

\bottomrule
\end{tabularx}
\label{tab:geometric_parameters}
\end{table}
\subsection{Precision grasp}
\begin{figure}[!ht]
    \centering
    \includegraphics[width=1\linewidth]{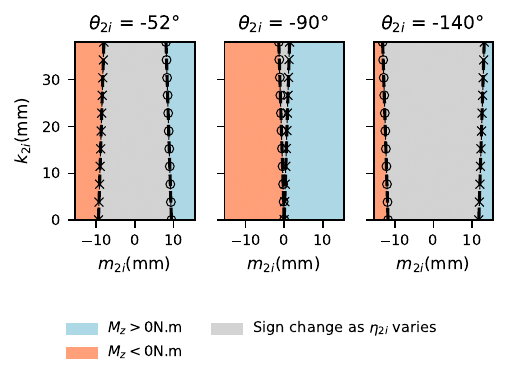}
    \caption{Sign of the moment around ${\bf z}_0$ as a function of the contact force location on the distal phalange}
    \label{fig:precisionstab}
\end{figure}

To study the stability of the grasp during a precision grip, the sign of $M_z$ is analyzed to ensure that the finger rotates in the correct direction, thereby preventing the object from being ejected. According to Eq.~\eqref{moment1}, the friction coefficient $\mu_{2i}$ has a negligible impact on the results, as the angle $\psi$ approaches 90°. Consequently, $\mu_{2i}$ is set to zero for this analysis. In contrast, the friction parameter $\eta_{2i}$ plays a more significant role. The equilibrium limit is defined by a cone, such that equilibrium occurs for values within the range $-\eta_{\text{static}} < \eta_{2i} < \eta_{\text{static}}$. The fingers are currently made of PLA, although the surface properties can be adjusted, while the objects to be grasped may consist of various materials (such as metal, fabric, or plastic). Therefore, an average value is adopted to ensure reasonably representative results with a static friction coefficient set to $\eta_{\text{static}}=0.3$. Figure~\ref{fig:precisionstab} shows the rotation direction as a function of $k_{2i}$ and $m_{2i}$ for different values of $\theta_{1i}$, with the direction indicated by color.

 The stability condition requires that, for any $\theta_{2i}$, $k_{2i}$, and $m_{2i}$, a stable contact position is maintained within the phalanx. This ensures that the grasp remains secure and the object is not at risk of being ejected due to contact forces acting outside the effective area of the phalanx. The finger tends to rotate to reach a stable position, $M_z=0 \text{N.m}$, along the phalanx. A condition on the distal phalanx width, $n_2$, can then be derived from Eq.~\eqref{moment1}, to ensure that the stable position always remains within the phalanx throughout the finger's range of motion:
\begin{equation}
    n_2\geq \frac{1}{s\psi-\mu_{static} c\psi} (l_2\eta_{static}c\psi+l_1\eta_{static}c\theta_{2imin}).
    \label{conditionwidth}
\end{equation}
\subsection{Power grasp}
For an enveloping grasp, it is essential to analyze the force distribution among the fingers to assess grasp stability as a function of the finger rotation.

According to Eq.~\eqref {forcecontact}, the contact forces depend on five parameters, which make the grasp-state space difficult to visualize. It can be observed that the contact forces are independent of $\theta_{1i}$ and $\theta_{2i}$, since these rotational parameters do not alter the kinematic configuration of the enveloping grasp. Consequently, the force distribution along the finger remains unaffected by variations in these two angles. According to Birglen \cite{birglen_design_2008}, $k_{1i}$, $\mu_{1i}$ and $\mu_{2i}$ have little influence, and are therefore omitted. 

We consider the case $k_{1i} = l_1/2$ and $k_{2i} = l_{2}/2$ as their influence is less significant than that of $m_{1i}$ and $m_{2i}$. 

\begin{figure}[!ht]
    \centering
    \includegraphics[width=0.9\linewidth]{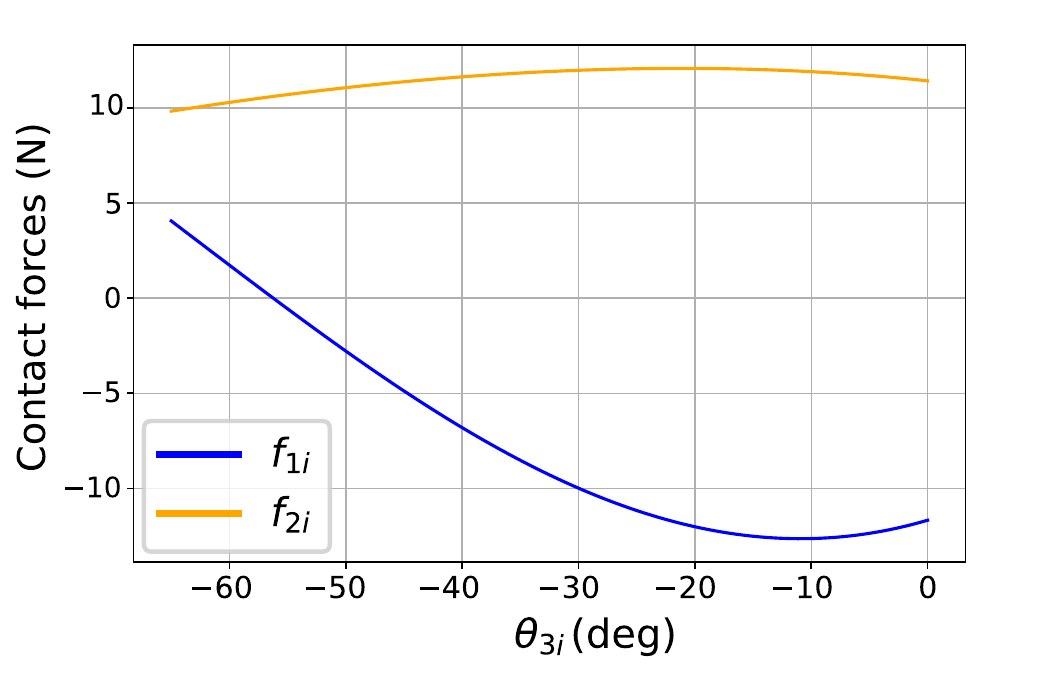}
    \caption{Contact forces as a function of $\theta_{3i}$}
    \label{fig:forces}
\end{figure}

As explained in Section~\ref{sec:power grasp}, a grasp is stable if and only if the contact forces between the object and the phalanges are positive. Figure~\ref{fig:forces}  illustrates these contact forces as a function of $\theta_{3i}$, considering a worst-case no-friction scenario for stability, as described in \cite{birglen_design_2008}. It can be observed that both contact forces are positive when $\theta_{3i}$ is less than $-56^\circ$.
Therefore, only values of $\theta_{3i}$ below this boundary are considered as they correspond to the only configurations that ensure a stable distribution of contact forces. 

 \begin{figure}[!ht]
    \centering
    \includegraphics[width=0.9\linewidth]{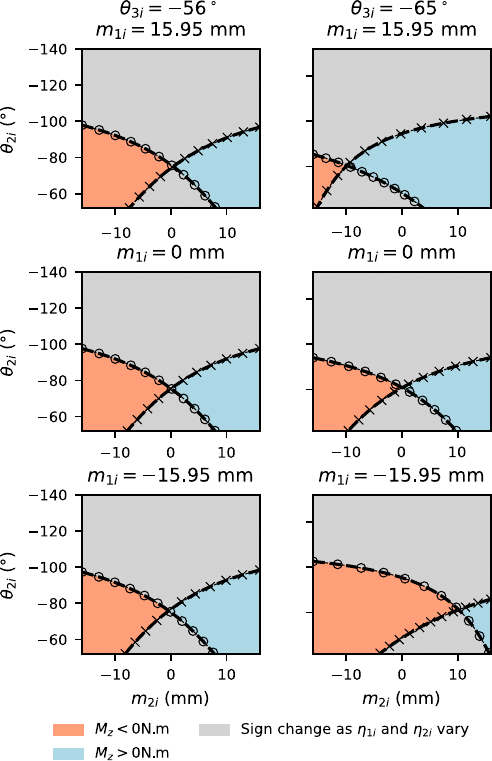}
    \caption{Sign of the moment around $\bf{z}_{0i}$ as a function of the contact forces locations on $\bf{x}_{1i}$ and angles $\theta_2$, $\theta_3$}
    \label{fig:momentenvel}
\end{figure}

Figure~\ref{fig:momentenvel} illustrates the influence of $m_{1i}$ and $m_{2i}$ as functions of $\theta_{2i}$ and $\theta_{3i}$. For different values of $m_{1i}$, the sign of the moment may vary with $\theta_{2i}$ and $\theta_{3i}$ for a fixed value of $m_{2i}$. 
When $m_{1i} = 15.95\ \text{mm}$, the curve $M_z = 0\text{N·m}$ is symmetric with respect to the case $m_{1i} = -15.95\ \text{mm}$, 
since the finger rotates in the same way for the pairs $(m_{1i}, m_{2i})$ and $(-m_{1i}, -m_{2i})$.
It can be observed that when $\theta_{3i} = -56^\circ$, the value of $m_{1i}$ does not influence the sign of the moment. This can be explained by the fact that, at $\theta_{3i} = -56^\circ$, $f_{1i} = 0\text{N}$. In contrast, for $\theta_{3i} = -65^\circ$, $m_{1i}$ has a more pronounced effect on the moment as $f_{1i}$ increases.
It can be observed that the grasp may become unstable for certain values of $m_{1i}$ and $m_{2i}$, depending on $\theta_{2i}$.
Indeed, variations in $\eta_{1i}$ and $\eta_{2i}$ can affect the sign of $M_z$, causing the location of the stable contact point to move outside the finger, which compromises the grasp stability. For values greater than $-100^\circ$, a stable contact position can be achieved along the phalanx. However, for smaller values, ejection may occur. This behavior can be explained by the fact that the distal phalanx moves closer to a horizontal orientation, reducing the ability to sustain a stable contact. These results could play a significant role in the design of the hand, particularly in determining its dimensions based on the minimum size of objects that must be grasped.
\subsection{Distribution of the actuation torque}
The differential mechanism distributes the actuation torque for the three fingers. Its parameters influence the hand’s adaptability. 

If the springs have infinite stiffness, it is as if there were no springs at all. In this case, the mechanism loses dexterity: if one finger is blocked, the others will be blocked as well. Furthermore, it can be seen that the distance (${A_{2i}A_{3i}}$) has a significant impact on the finger’s actuation torque: the larger the distance, the greater the actuation torque.

Furthermore, force control is limited in this type of spring-differential mechanism. Indeed, it is not possible to specify an exact force for each output, since the output forces are coupled. If one finger is blocked, the resulting redistribution of forces affects the other outputs, which are directly related to the actuation torques of the fingers. 
\subsection{Experimental result}

\begin{figure} [!ht]
    \centering
    \includegraphics[width=1\linewidth]{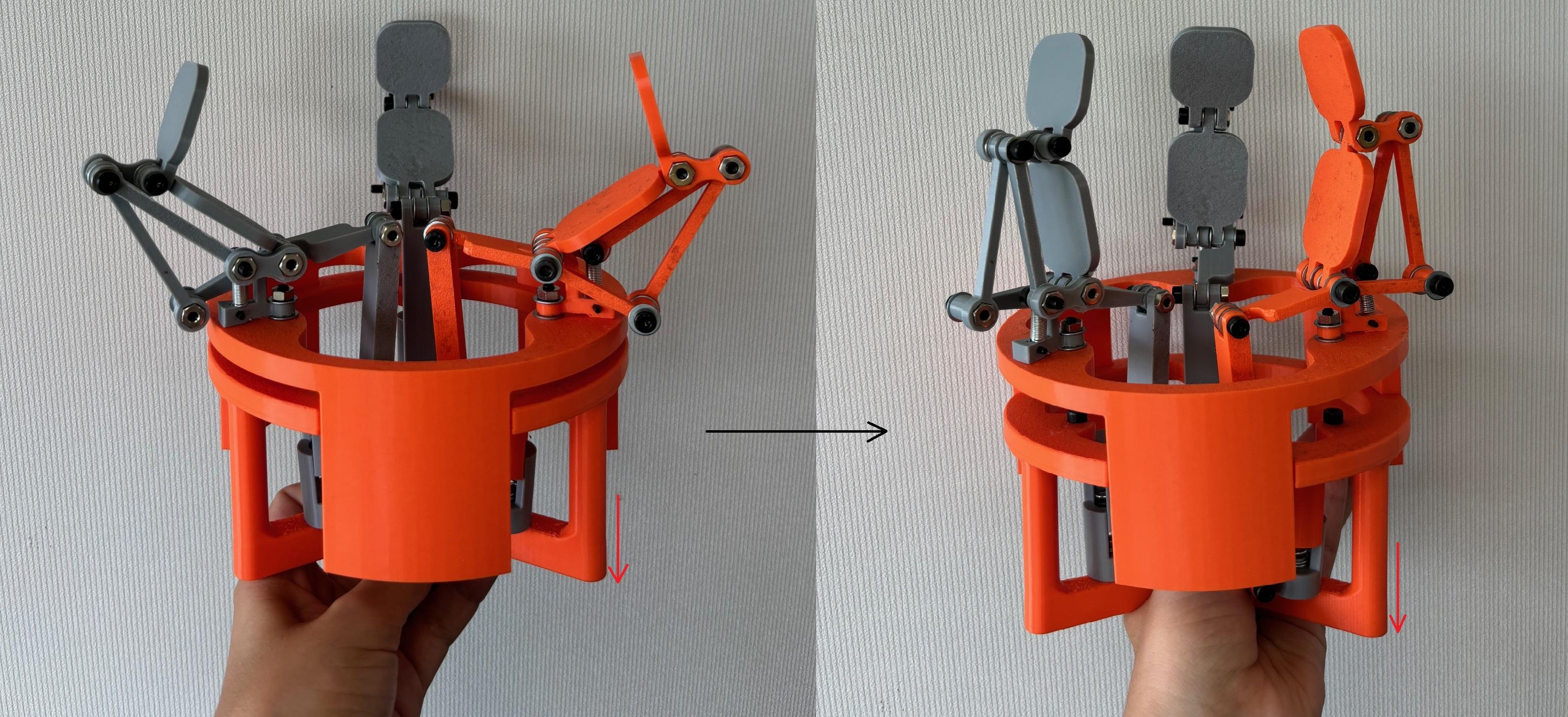}
    \caption{Hand Prototype manually actuated thanks to a slider: the fingers close when the slider is pulled.}
    \label{fig:proto}
\end{figure}

\textcolor{black}{A physical prototype was designed to satisfy the phalanx width stability condition defined in Eq.~\eqref{conditionwidth}.} A fully mechanical hand prototype was fabricated using 3D printing. Its operation relies on the translation of the lower part, actuated manually by the user through a pulling motion (Fig.~\ref{fig:proto}).

To test different grasping modes and their stability, several objects were selected: a pencil holder, a wallet, a small box, and a bottle. These object shapes enable four classic grasping types: precision-spherical grasp, precision-cylindrical grasp, enveloping-spherical grasp, and enveloping-cylindrical grasp. Table~\ref{tab:grasps} illustrates these grasping modes successfully executed with the prototype. \textcolor{black}{These experiments demonstrate promising preliminary results. An actuated prototype is currently under development to enable a deeper comparative analysis with state-of-the-art robotic hands.}

\begin{table}[!ht]
\caption{Types of grasps illustrated for a pencil holder, wallet, small box, and bottle}
\centering
\renewcommand{\arraystretch}{1.5} 
\begin{tabular}{c c c}
\toprule
 & Spherical grasp & Cylindrical grasp \\
\midrule
Precision grasp & \includegraphics[width=0.3\linewidth, angle=180]{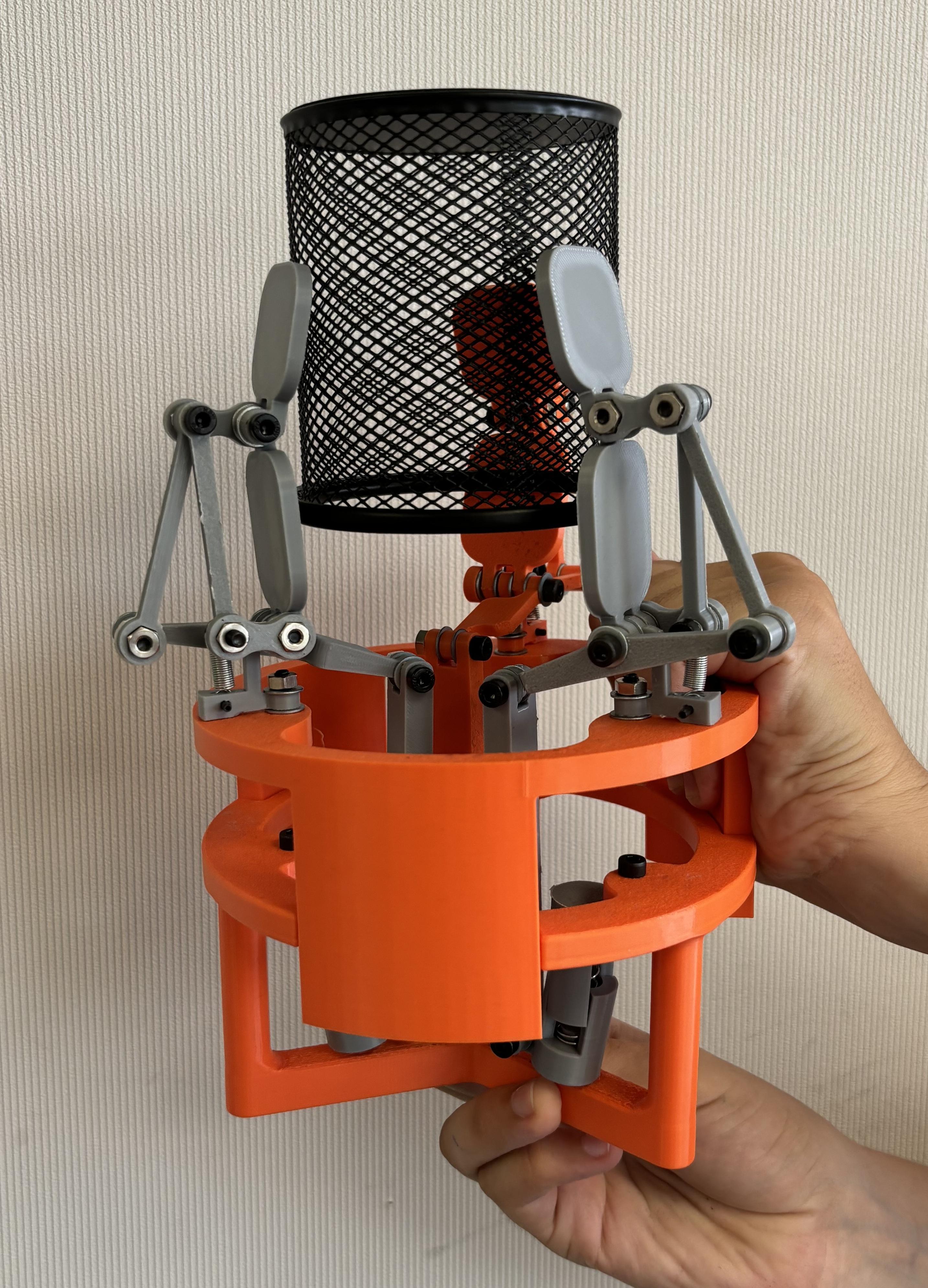} & \includegraphics[width=0.245\linewidth, angle=180]{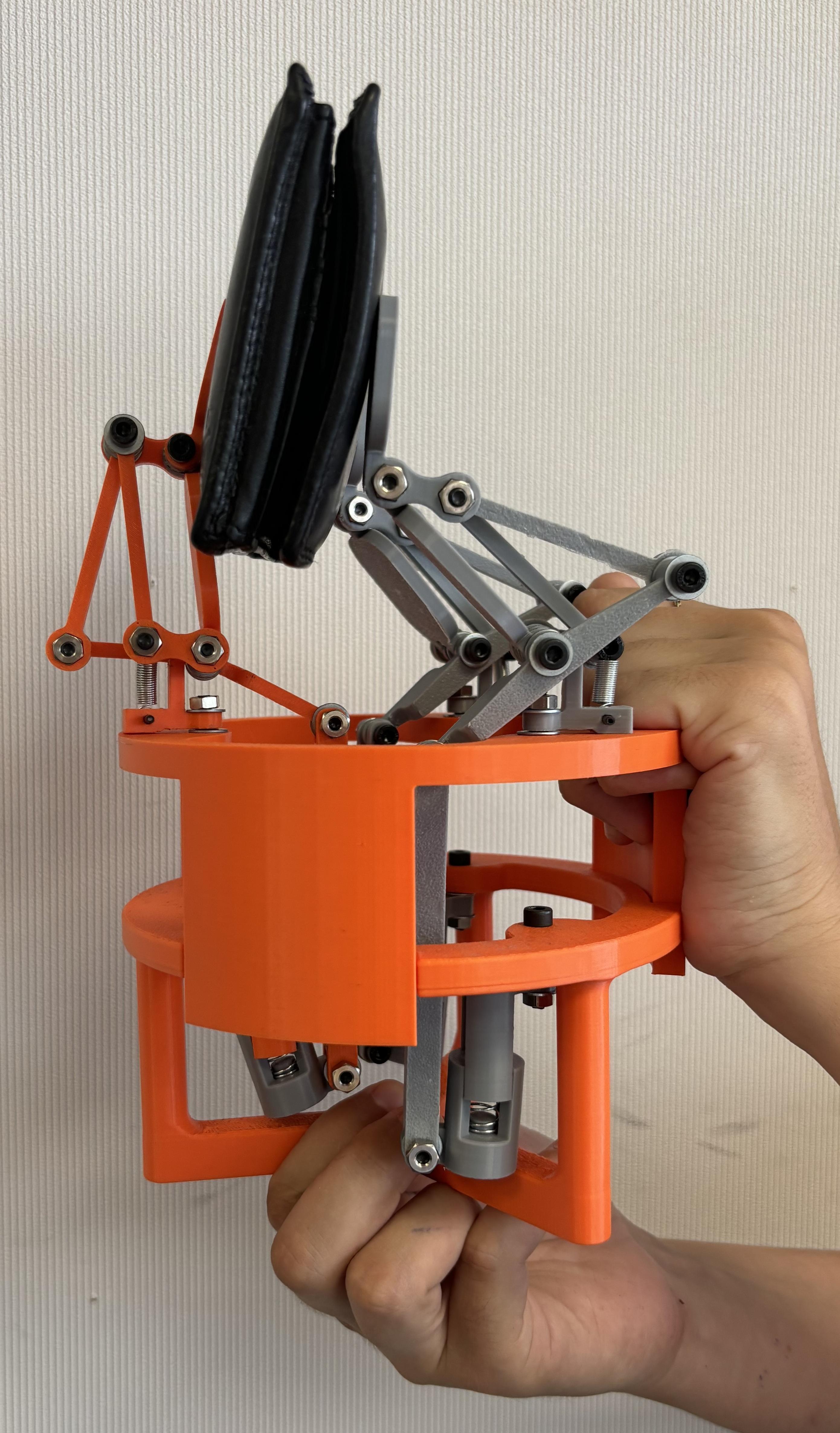} \\
Enveloping grasp & \includegraphics[width=0.3\linewidth, angle=180]{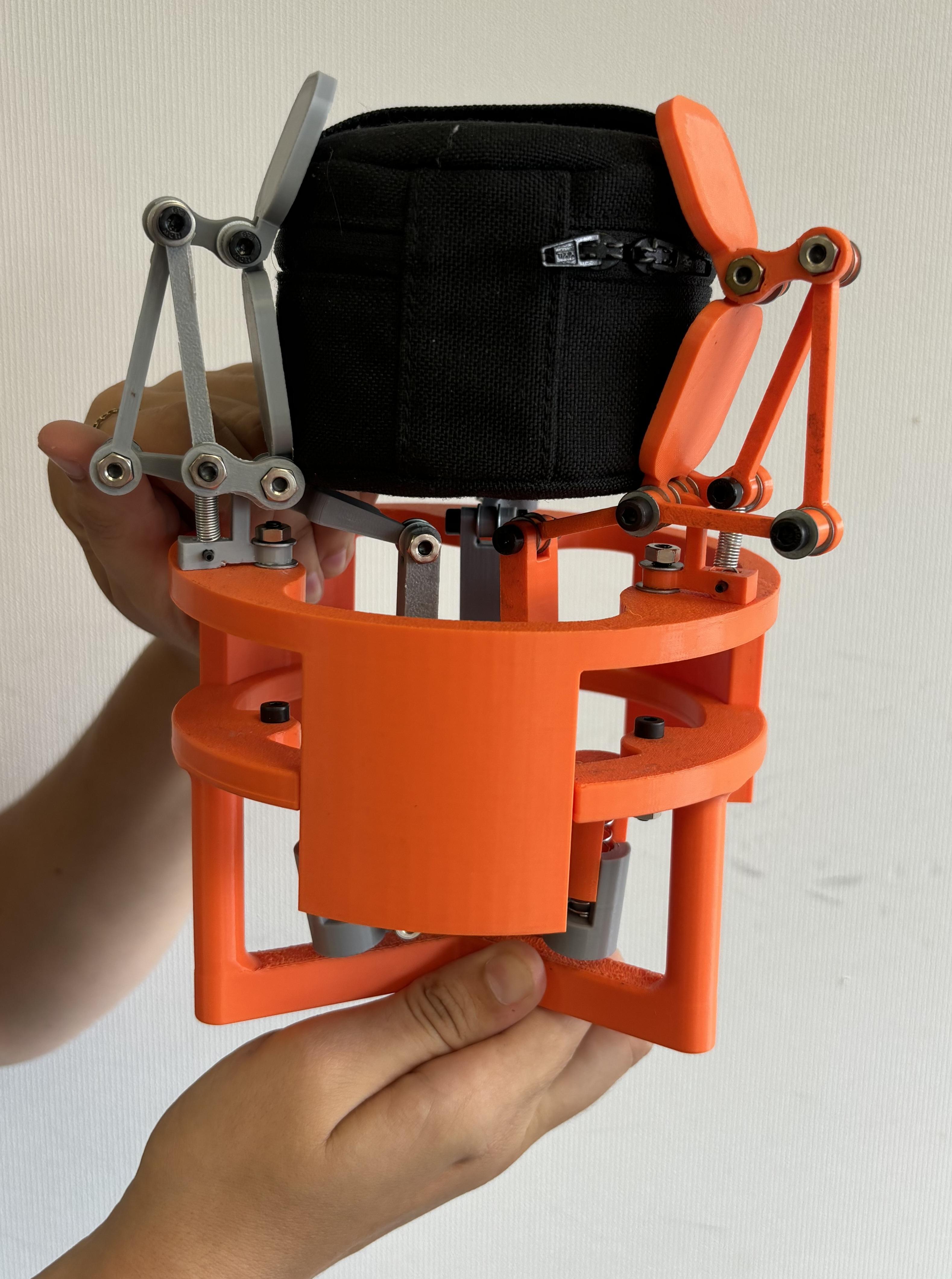} & \includegraphics[width=0.305\linewidth, angle=180]{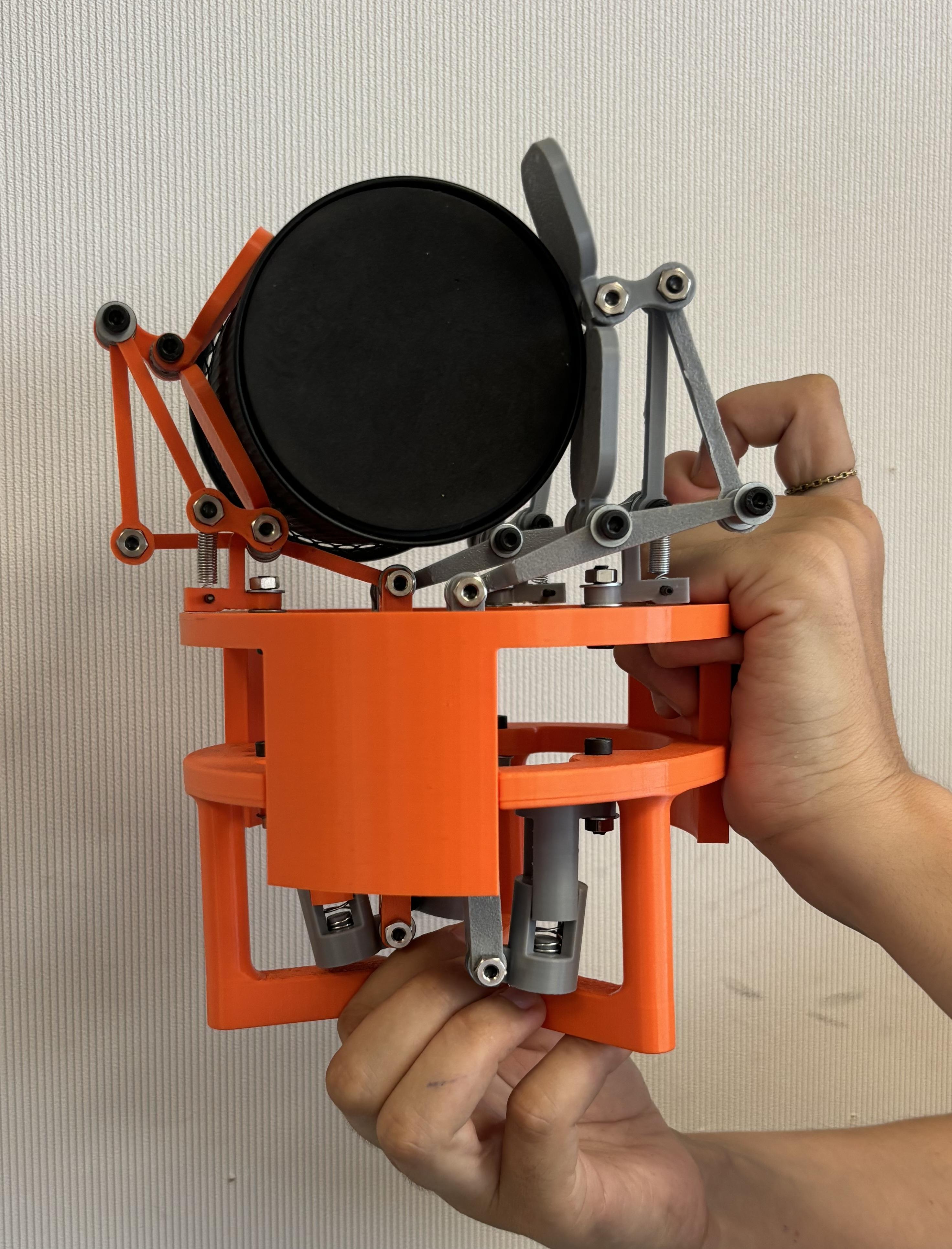} \\
\bottomrule
\end{tabular}
\label{tab:grasps}
\end{table}
\section{Discussion}
These initial results are encouraging, as the prototype meets the requirements regarding grasp diversity and stability. However, several simplifications have been made, and some aspects have been neglected, which should be taken into account in future work or at least be the subject of a sensitivity analysis.

In the theoretical models, the phalanx thickness was neglected. In future work, both the thickness and shape of the phalanx, which was assumed to be flat in this work, will be taken into account. \textcolor{black}{The thickness would have only a limited impact on precision grasp, but for enveloping grasps, it could influence finger rotation.} The study of a curved shape, similar to that of a human finger, for example, could be particularly interesting for studying the passive rotation of the finger and could also affect its stability.

Regarding the influence of the springs, the effect of $S_{2i}$ has not been taken into account in either stability analysis. In the case of parallel grasps, this spring is not expected to affect the results as it remains unstretched. For enveloping grasps, Birglen \cite{birglen_design_2008} demonstrates that its influence remains limited, slightly improving overall stability. In addition, the springs $S_{3i}$ and $S_{4i}$ have also been neglected. These elements, however, constrain the finger’s rotation and tend to bring it back to its initial position. Therefore, determining an appropriate trade-off in their stiffness is essential to ensure a proper return to the initial configuration while minimizing their impact on finger rotation.

Regarding the modeling of contact conditions during enveloping grasp, another limitation lies in the assumption that the initial position of the contact forces is fixed at the middle length of each phalanx. Furthermore, the movements and changes in orientation that occur during grasping have not been taken into account. Although this approach provides a first approximation, further developments are needed to achieve a more accurate representation of hand behavior during grasping.

Finally, from an experimental point of view, the 3D-printed hand prototype has provided initial results regarding the grasping of objects of various shapes. Future work will focus on developing a specific experimental protocol and designing a new instrumented robotic hand. The aim is to conduct a more comprehensive study of the hand and the grasp stability, using extensive experimental studies to rigorously evaluate theoretical models.
\section{Conclusions}
This work presents a simplified and innovative approach to designing robotic fingers that can perform multiple grasp types with both stability and adaptability. The proposed architecture allows for transitions between configurations by integrating parallel and self-adaptive grasping modes, enabling the hand to adapt to a variety of object shapes.

Theoretical analysis has demonstrated the conditions under which stable contact is maintained, particularly in precision and enveloping grasps. The relevance of the kinematic model has been validated using an initial 3D-printed prototype showing the adaptability of the fingers and the stability of the grasp.

The use of underactuation, both within and between fingers, has proven to be effective in simplifying mechanical design while preserving functional versatility. This strategy reduces the number of actuators required and enhances passive adaptability, making the system well-suited for industrial and medical applications where reliability and cost-efficiency are critical.

This architecture opens promising avenues for applications where adaptability, simplicity, and robustness are essential. It is particularly well-suited for agricultural robotics, where delicate and irregular objects such as fruits must be grasped reliably. In logistics and warehouse automation, the ability to handle diverse object shapes with minimal control effort offers clear advantages. The design also holds potential for assistive technologies and prosthetics, providing lightweight and compliant grasping mechanisms for daily use. Additionally, its adaptability makes it relevant for waste sorting and recycling tasks, where unpredictable object geometries demand flexible and resilient handling strategies.

However, the current prototype relies on manual actuation, which limits repeatability and control precision. Future work will focus on integrating an electric motor to automate the hand motion and enable controlled approach strategies to improve grasp stability and precision.

Ultimately, this study lays the groundwork for the development of compliant robotic hands that combine mechanical simplicity with functional richness, opening new possibilities for autonomous manipulation in unstructured environments.

\section*{Acknowledgment}
This work was supported by funding from the French government, managed by the National Research Agency under the France 2030 program, reference ANR-22-EXOD-0003, within the PEPR Organic Robotics.
\bibliographystyle{asmeconf}  
\bibliography{These_Leonie}

\end{document}